%% file: main.tex
\documentclass[10pt,twocolumn,letterpaper]{article}

 \usepackage{iccv}              


\definecolor{iccvblue}{rgb}{0.21,0.49,0.74}

\usepackage[pagebackref,breaklinks,colorlinks,allcolors=iccvblue]{hyperref}
\usepackage{multirow}
\usepackage{marvosym}   
\def\paperID{5} 
\def\confName{ICCV}
\def\confYear{2025}

\title{Dynamic Inter-Class Confusion-Aware Encoder for Audio-Visual Fusion in Human Activity Recognition}
\makeatletter
\def\@fnsymbol#1{%
  \ifcase#1 \or \Letter \else\@ctrerr\fi
}
\author{%
  Kaixuan Cong$^{\spadesuit}$, 
  Yifan Wang$^{\heartsuit}$,
  Rongkun Xue$^{\spadesuit}$,
  Yuyang Jiang$^{\heartsuit}$, 
  Yiming Feng$^{\spadesuit}$,
    Jing Yang$^{\spadesuit}$%
    \thanks{Corresponding author.\\
    Email: \href{mailto:20011004@stu.xjtu.edu.cn}{20011004@stu.xjtu.edu.cn}
    ; \href{mailto:jasmine1976@xjtu.edu.cn}{jasmine1976@xjtu.edu.cn}.}\\
  $^{\spadesuit}$School of Automation Science and Engineering, Xi'an Jiaotong University, Xi'an, China\\
  $^{\heartsuit}$School of Software Engineering, Xi'an Jiaotong University, Xi'an, China
}
\begin{document}
\maketitle
\input{sec/0_abstract}    
\input{sec/1_intro}
\input{sec/3_method}
\input{sec/4_experiment2}
\input{sec/5_conclusion}
{
    \small
     \bibliographystyle{ieeenat_fullname}
     \bibliography{main}
}

\end{document}

%% file: sec/0_abstract.tex
\begin{abstract}   
Humans do not understand individual events in isolation; rather, they generalize concepts within classes and compare them to others. Existing audio-video pre-training paradigms only focus on the alignment of the overall audio-video modalities, without considering the reinforcement of distinguishing easily confused classes through cognitive induction and contrast during training. This paper proposes the Dynamic Inter-Class Confusion-Aware Encoder (\textit{DICCAE}), an encoder that aligns audio-video representations at a fine-grained, category-level. \textit{DICCAE} addresses category confusion by dynamically adjusting the confusion loss based on inter-class confusion degrees, thereby enhancing the model's ability to distinguish between similar activities. To further extend the application of \textit{DICCAE}, we also introduce a novel training framework that incorporates both audio and video modalities, as well as their fusion. To mitigate the scarcity of audio-video data in the human activity recognition task, we propose a cluster-guided audio-video self-supervised pre-training strategy for \textit{DICCAE}. 
\textit{DICCAE} achieves near state-of-the-art performance on the VGGSound dataset, with a top-1 accuracy of 65.5\%. We further evaluate its feature representation quality through extensive ablation studies, validating the necessity of each module.

\end{abstract}

%% file: sec/1_intro.tex
\section{Introduction}
\label{sec:intro}

In the field of multimodal representation learning, both unsupervised and supervised methods—such as SimCLR~\cite{chen2020simple}, GCN~\cite{Kipf17}, and Swin Transformer~\cite{liu2021swin}—aim to progressively organize unstructured information into distribution centers. As the number of categories and data dimensions increases, the information density grows, making it more challenging for models to distinguish different features. This drives the development of deeper architectures~\cite{he2016deep} and more specialized networks tailored to specific modalities~\cite{xu2023multimodal}.
\begin{figure}[t]
  \centering
  \includegraphics[width=0.5\textwidth]{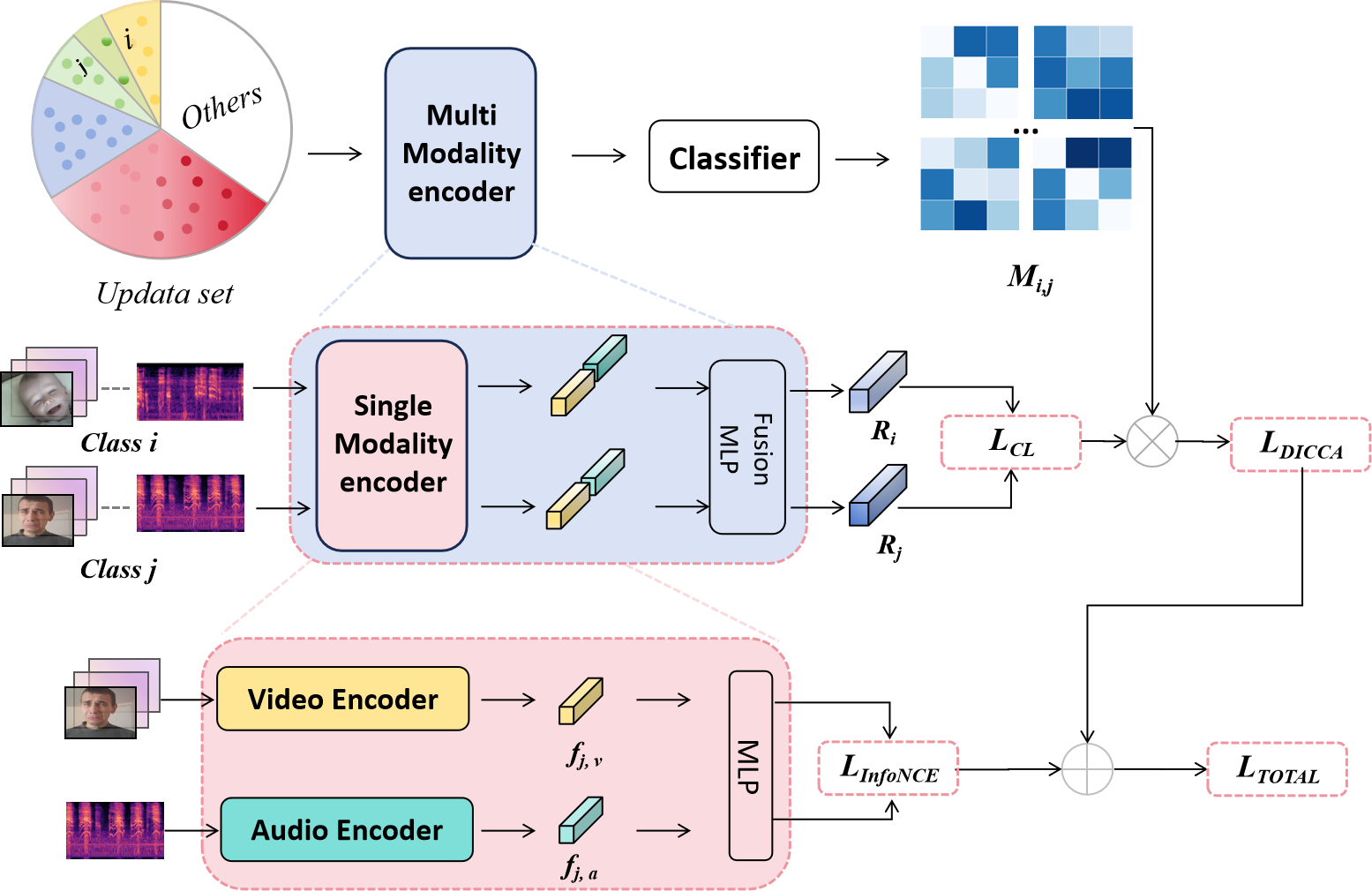} 
  \caption{The diagram illustrates Dynamic Inter-Class Confusion-Aware Encoder (\textit{DICCAE}). The first layer extracts features from audio and video inputs, which are processed by separate encoders. The second layer computes the  \( \mathcal{L}_{CL} \) based on fused audio-video features. The third layer generates the confusion matrix \( \mathcal{M}_{ij} \) and then computes the loss.}
  \label{fig:training-architecture}
\end{figure}
Compared to static images, audio and video have higher information density, which drives researchers to develop larger models~\cite{zhang2019multimodal}. However, this paradigm is not universally applicable to multimodal tasks. 

Human activity recognition is one of the core tasks that simultaneously involves both audio and video information. Although models such as ViT~\cite{Dosovitskiy20}, AST~\cite{gong2021ast}, and Whisper~\cite{Radford23} have made significant advances in single-modal activity recognition tasks, there are inherent challenges when dealing with high-density information in audio and video, particularly in their respective modality representations. Specifically, many similar activity categories are difficult to distinguish accurately when recognized using only a single audio or video modality (e.g., 'fall/lie down' and 'cough/sneeze'). These categories inherently exhibit confusion when analyzed in a single modality, but from a multimodal perspective, audio and video offer complementary information for these categories~\cite{shaikh2024multimodal}.Properly fusing audio and video information is the optimal solution for distinguishing easily confused activity categories. Given the differing information distributions and network architectures across audio and video modalities, building a unified framework with improved cross-modal learning capabilities has become a key research focus~\cite{gong2022uavm,pramanik2019omninet, srivastava2024omnivec, Baltrusaitis19, Akbari21, Jaegle21, Alayrac22, Zellers21, Wang23}.

The aforementioned works focus on self-supervised alignment of audio and video features during the pre-training phase to obtain better feature representations for downstream tasks. However, these models do not account for a key factor in the construction of the human cognitive system: induction and comparison. 'Humans do not understand individual events in isolation; instead, they generalize concepts within classes and compare them to others.' These models align audio and video modalities at a coarse level by reducing the spatial distance between their features, without incorporating a design that guides the model to perform induction and comparison at the fine-grained category level. As a result, they fail to leverage the complementary information between audio and video to distinguish easily confused categories effectively.

How to guide the model to distinguish easily confused categories during training, research in ~\cite{wu2016multi} improves video classification performance through a loss function designed for multi-stream, multi-class fusion, while ~\cite{khosla2020supervised} enhances the model's ability to distinguish similar categories by designing contrastive loss and network structures for better representation learning. However, these methods are implemented in a single modality. These methods often rely on static priors or sampling strategies to handle class boundaries, which do not adapt dynamically during training. This overlooks the dynamic nature of the cognitive system and the fact that the degree of confusion between different classes varies.

Therefore, a training framework for audio-video fusion that can mimic human cognitive induction and comparison, while dynamically adjusting the model’s focus on confused classes by quantifying the confusion degree between different classes, is urgently needed. This paper proposes the Dynamic Inter-Class Confusion-Aware Encoder(\textit{DICCAE}), an encoder designed to align audio-video features at a fine-grained level across different categories. The framework consists of two main components: firstly, the Dynamic Inter-Class Confusion-Aware loss, which introduces a confusion loss for comparing fused audio-video features at the category sample level, and a method for quantifying the degree of confusion between classes based on their overlap in feature space. Secondly, a joint audio-video training framework is proposed, enabling the model to align features between audio, video, and categories.

However, another significant challenge in the field of audio-video representation is the scarcity of high-quality labeled audio-video datasets, especially for human activity recognition. The only large-scale high-quality publicly available datasets for this task are Audioset~\cite{gemmeke2017audio}, VGGSound~\cite{chen2020vggsound}, EPIC-KITCHENS-100~\cite{damen2018scaling}, and UCF101~\cite{soomro2012ucf101}, with most others being small-scale. However, the labels used to guide the model in learning complex information distributions do not necessarily need to align perfectly with human cognition. Self-supervised pre-training guided by clustering results has also shown great efficacy~\cite{hsu2021hubert}. The framework proposed in this paper, which aligns audio-video fusion features at a fine-grained category level, inherently benefits from such a clustering-guided self-supervised training paradigm, as it can obtain category-level fusion features during pre-training. Therefore, this paper proposes a cluster-guided audio-video self-supervised strategy. Specifically, we use a pre-trained audio-video encoder to process datasets and obtain concatenated audio-video features, using clustering results as category labels to guide the model in fine-grained audio-video feature alignment through contrast and induction.

The contributions can be summarized as follows:

\begin{itemize}
\item We propose the Dynamic Inter-Class Confusion-Aware Encoder, designed to reduce category confusion in multimodal Human Activity Recognition tasks, particularly for complex spatio-temporal data such as audio and video.

\item We introduce the concept of inter-class confusion, provide a quantitative metric to evaluate the confusion, and use it to dynamically adjust the loss weight during training, thereby focusing on classes with high confusion.

\item We propose a cluster-guided self-supervised training strategy for audio-video fusion, enabling the model to use unlabeled fused audio-video data as the training set in the self-supervised phase, thereby aligning audio-video features at a fine-grained level.

\item We introduce an end-to-end training framework and strategy for (\textit{DICCAE}). Our model achieves a top-1 accuracy of 60.1\% in audio-only classification and 65.5\% in audio-visual classification on the VGGSound dataset. A series of ablation experiments demonstrate the effectiveness of our framework.
\end{itemize}

%% file: sec/3_method.tex
\section{Method}
\label{sec:method}

As shown in \cref{fig:training-architecture}, to tackle category confusion in Human Activity, we propose the Dynamic Inter-Class Confusion-Aware encoder (\textit{DICCAE}). This method incorporates a confusion loss to mimic human cognition—bringing similar samples closer while pushing dissimilar ones apart—and an end-to-end training framework and strategy for audio–visual data. For the confusion loss, It dynamically weights the confusion loss based on this confusion, improving the model’s focus on difficult-to-distinguish categories. We also present a training framework that applies this loss to audio-video fusion scenarios. To address the scarcity of multimodal samples for Human Activity recognition, we propose utilizing multiple unlabeled datasets during the pre-training phase, where the clustering results obtained through the K-means algorithm are used as category labels to guide the model in learning fine-grained audio-video fusion features.

\subsection{Dynamic Inter-Class Confusion-Aware Loss}

This loss simulates human cognition by incorporating and quantifying inter-class confusion. The design of loss of \textit{DICCAE} is outlined in the following steps:
\subsubsection{Confusion Loss}

Confusion loss simulates the contrast and induction process in human cognition. The training framework takes pairs of data categories \( (x_i, x_j) \), where \( x \) can be audio or video. These pairs are passed through feature extractors to obtain hidden space features \( (f_i, f_j) \) for each category.

The confusion loss \( \mathcal{L}_{CL} \) is calculated based on \( (f_i, f_j) \), as shown in Formula 1:
\[
\mathcal{L}_{CL} = -y \log F_p(f_i, f_j) - (1 - y) \log(1 - F_p(f_i, f_j))\tag{1}
\]
where \( y = 1 \) if \( i \) and \( j \) are the same category, and \( y = 0 \) otherwise. Here, \( F_p(f_i, f_j) \) computes the similarity between the features, and the loss encourages features of the same category to be more similar, while pushing features of different categories apart.

\subsubsection{Inter-Class Confusion Degree Calculation}

To quantify the inter-class confusion, we follow the steps outlined below:

1. Find the Centroid of Each Class:
For each class \( c \), we compute the centroid by averaging the hidden space features of all samples in that class. This centroid represents the point that minimizes the distances to all other points within the class.

2. Calculate the Coverage Range of Each Class:
   For each class \( c \), we fit an enclosing circle that covers 95\% of the feature points in that class to avoid interference from outliers. The circle is calculated using principal component analysis followed by circle fitting. The goal is to ensure that the circle covers the central 95\% of the samples for each class.

3. Calculate Inter-Class Confusion Degree:
   For each pair of classes \( i \) and \( j \), we calculate the overlap between their respective minimal enclosing circles. Let \( r_i \) and \( r_j \) be the radii of the enclosing circles for classes \( i \) and \( j \), respectively, and \( d_{ij} \) be the distance between their centers. The inter-class confusion degree \( \mathcal{M}_{ij} \) is computed as:

   \[
   \mathcal{M}_{ij} = \frac{\max(0, r_i + r_j - d_{ij})}{d_{ij}}\tag{2}
   \]

   This formula measures the overlap between the circles and the relative distance between their centers. A higher value indicates greater overlap and, thus, higher confusion between the two classes.

4. Feature Quality Evaluation:
   The distribution of inter-class confusion degrees is used to evaluate the quality of the feature representation. If the confusion degree distribution is centered around zero and has low variance, the feature quality is considered good, meaning the model has effectively separated the classes.

\subsubsection{Dynamic Weighting}

To improve the model's focus on easily confused categories, we introduce a dynamic weighting mechanism based on the inter-class confusion degree. The procedure is as follows:

1. Selection of the Update Set:
   At the end of each training epoch, we select a portion of the test set as the update set, which is used to calculate the inter-class confusion degree matrix.

2. Calculation of the Inter-Class Confusion Degree Matrix:
   After each epoch, the model is used to infer the update set, and a \( C \times C \) inter-class confusion degree matrix \( \mathcal{M} \) is constructed. Each element \( \mathcal{M}_{ij} \) represents the confusion degree between classes \( i \) and \( j \).

3. Normalization:
   The values of the confusion degree matrix are normalized to the range \( [0, 2] \) using the following formula:

   \[
   \hat{\mathcal{M}}_{ij} = \frac{\mathcal{M}_{ij} - \min(\mathcal{M})}{\max(\mathcal{M}) - \min(\mathcal{M})} \times 2\tag{3}
   \]

   This ensures that the confusion degree values are within a reasonable range for training.

4. Application of Dynamic Weighting:
   During training, for each pair of categories \( (i, j) \), the value \( \hat{\mathcal{M}}_{ij} \) from the confusion degree matrix is used as a dynamic weight for the confusion loss. The weighted confusion loss is calculated as:

   \[
   \mathcal{L}_{\textit{DICCAE}} = \sum_{i,j} \hat{\mathcal{M}}_{ij} \cdot \mathcal{L}_{CL}(i,j)\tag{4}
   \]

   where \( \mathcal{L}_{CL} \) is the confusion loss  between categories \( i \) and \( j \). By applying dynamic weighting, the model focuses more on pairs of categories that are difficult to distinguish, thereby improving the model's performance in distinguishing similar categories.

\subsection{Audio-Video fusion training framework}

As shown in \ref{fig:training-architecture}, the \textit{DICCAE} in audio-video fusion training requires a two-layer framework, which from bottom to top corresponds to cross-modal feature alignment, and inter-class loss calculation.

The first layer focuses on processing audio and video separately using encoders for each modality. These encoders produce low-level feature representations, which are then used to calculate the InfoNCE loss, enabling the model to contrast features within each modality.

The second layer combines the features from both modalities for each class pair, aligning the audio-video representations in class level. The combined features are passed through an MLP, which computes the LCF loss and the classification loss, ensuring better separation between similar activity categories.

Additionally, during training, a portion of the test set is separated as the update set for calculating the inter-class confusion degree matrix, which is used to dynamically adjust the weight, emphasizing the more challenging-to-differentiate classes. The total training loss is a combination of the classification losses and the \textit{DICCAE} loss, facilitating effective learning for multimodal fusion.

This architecture provides a comprehensive approach for managing category confusion and enhancing feature alignment in audio-visual learning tasks.

\subsection{Cluster-guided Audio-Video Self-supervised Strategy}

To effectively address the issue of sparse labeled data, we propose a cluster-guided audio-video self-supervised strategy. This strategy utilizes clustering on a large-scale unlabeled dataset, using the K-means algorithm for self-supervised learning, which improves the model's understanding of multimodal data.

Specifically, we merge multiple Human Activity Recognition datasets containing audio and video information into a larger-scale unlabeled training set. To obtain the category label, we perform clustering using the K-means method. The settings for clustering are as follows:

\begin{enumerate}
    \item \textbf{Random Initialization}: Due to the sensitivity of the K-means algorithm to the initial cluster centers, we perform 20 random initializations of the K-means algorithm and select the optimal clustering result.

    \item \textbf{Cluster Label Assignment}: After clustering, each cluster is assigned a corresponding label, which represents the category information of the cluster center. During the training, these cluster labels will serve as self-supervised learning signals to guide the model in learning.

    \item \textbf{Refinement of Cluster}: Every 10 training epochs, we reapply K-means clustering using the DICCAE model’s audio-video fusion features.
\end{enumerate}

Through this cluster-guided self-supervised strategy, the model is able to acquire more fine-grained multimodal feature representations with the assistance of unlabeled data.

%% file: sec/4_experiment2.tex
\section{Experiments}
\label{sec:experiments}

\subsection{Dataset \& Training Configuration}
For the selection of pre-training datasets, we combine Audioset~\cite{gemmeke2017audio}, VGGSound~\cite{chen2020vggsound}, EPIC-KITCHENS-100~\cite{damen2018scaling}, and UCF101~\cite{soomro2012ucf101} into a larger dataset of approximately 2.3 million samples. Regarding the choice of the K-value for K-means, we set K=300 by default in this experiment and update the clustering labels every 10 epochs.
All models were trained using the Adam optimizer with a learning rate of $1 \times 10^{-4}$ and a batch size of 32. The self-supervised phase ran for a maximum of 500 iterations, while the supervised phase was limited to 30 steps. A learning rate decay strategy was applied throughout training to ensure stable convergence.

\subsection{Experimental Setup}
To validate the effectiveness of the proposed \textit{DICCAE} for the multimodal Human Activity Recognition task, fine-tuning experiments were conducted on the VGGSound dataset. We evaluated the performance of DICCAE in both single-modal and multi-modal settings, analyzing its impact on feature representations through classification accuracy and inter-class confusion metrics.


\subsection{Impact of \textit{DICCAE} on Confusion Distribution}
DICCAE can optimize the distribution of inter-class confusion degrees in the model on a given dataset. The inter-class confusion degree distribution serves as a critical metric for evaluating the model's feature representation quality.

\begin{figure}[htbp]
  \centering
  \includegraphics[width=0.4\textwidth]{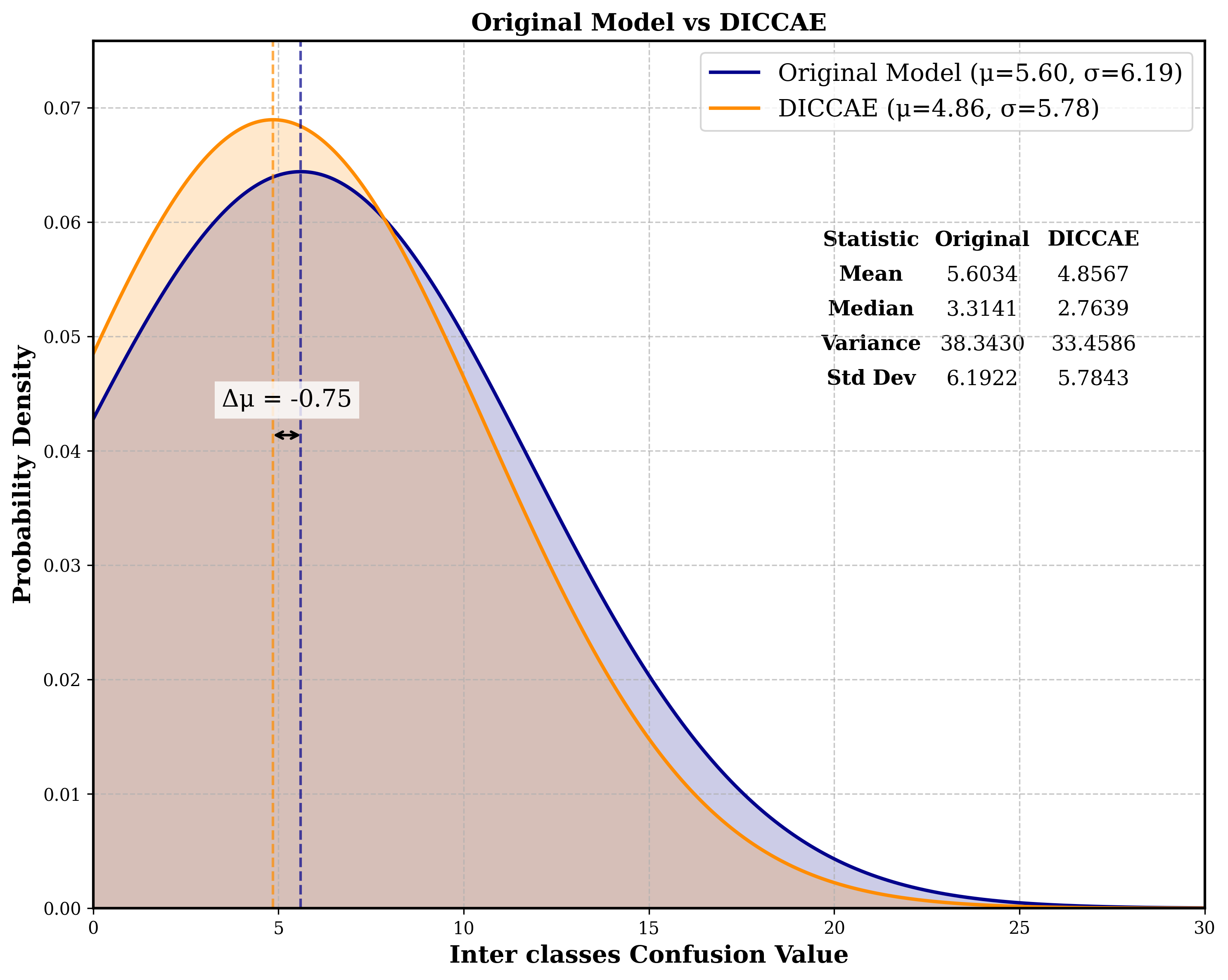} 
  \caption{Comparison of inter-class confusion degree distribution with and without \textit{DICCAE} on the VGGSound dataset.}
  \label{fig:confusion}
\end{figure}

Figure \ref{fig:confusion} illustrates a comparison of the inter-class confusion degree distributions, with and without \textit{DICCAE}, on the VGGSound dataset. Thanks to the application of \textit{DICCAE}, the mean of the distribution decreased by 0.75, and the variance reduced by 4.89, indicating that the feature representations produced by \textit{DICCAE} have less overlap between similar classes.

\subsection{Impact of \textit{DICCAE} on Classification Result}
\begin{figure}[htbp]
  \centering
  \includegraphics[width=0.48\textwidth]{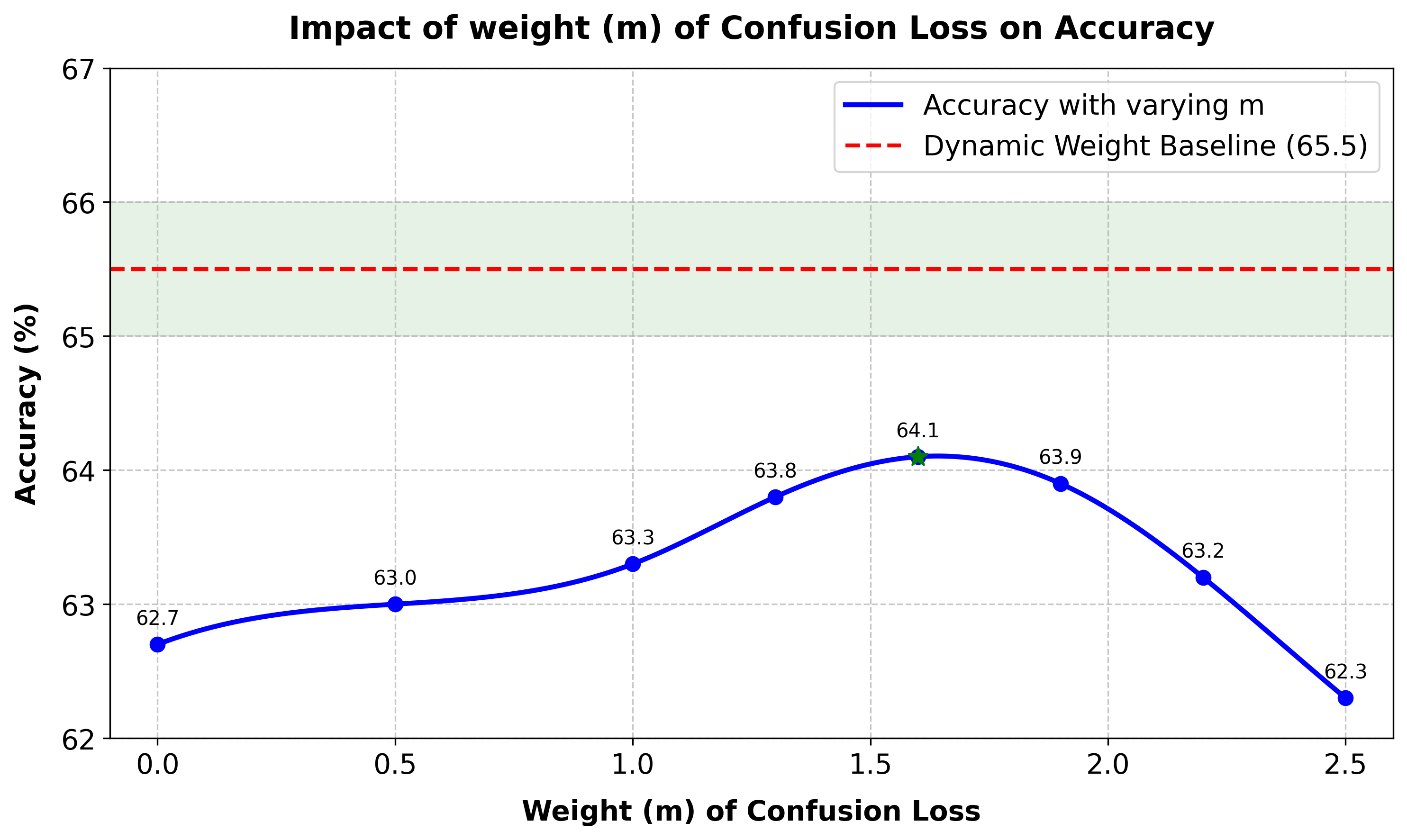} 
  \caption{Audio classification accuracy as a function of confusion loss weight.}
  \label{fig:dynamic-weighting}
\end{figure}
The experiment is conducted on the VGGSound dataset, and the evaluation metric is Top-1 accuracy.

First, we perform audio-video joint fine-tuning training on the VGGSound dataset and evaluate the accuracy, achieving results close to the state-of-the-art (SOTA).

The results in Table \ref{tab:vggsound-performance} demonstrate that \textit{DICCAE} is an effective method. Although applied to a standard audio-video contrastive learning framework, the encoder still achieves competitive performance 65.5\% even without additional optimization.

Next, we observe the model's classification performance by adjusting the weights of the confusion loss components, validating the effectiveness of dynamic weighting.

Figure \ref{fig:dynamic-weighting} shows the effect of fixed weights versus dynamic weighting based on inter-class confusion on training. The blue line represents accuracy with different fixed weights, while the green dashed line shows results with dynamic weighting. When the fixed weight exceeds a threshold, performance drops, this due to the model focusing too much on contrast and induction. In contrast, dynamic weighting yields more stable and better results.

\begin{table}[htbp]
\centering
\caption{Performance Comparison of Different Models}
\label{tab:vggsound-performance}
\begin{tabular}{lccc}
  \hline
  \multirow{2}{*}{\textbf{Method}} 
    & \multicolumn{3}{c}{\textbf{VGGSound (Acc.\%)}} \\
  \cline{2-4}
    & \textbf{A} & \textbf{V} & \textbf{A-V} \\ 
  \hline
  GBlend \cite{wang2020makes}       & $43.2$ & –    & –    \\
  Perceiver \cite{Jaegle21}         & –    & –    & 58.9 \\
  Attn AV \cite{fayek2020large}     & –    & –    & 60.5 \\
  MBT \cite{nagrani2021attention}   & $52.3$ & $51.2$ & $61.3$ \\
  CAV-MAE \cite{gong2022contrastive}& $59.5$ & $47.0$ & $65.5$ \\
  AudiovisualMAE \cite{Georgescu23} & $57.2$ & $50.3$ & $65.0$ \\
  DICCAE (ours)                     & $60.1$ & $52.3$ & $65.5$ \\
  \hline
\end{tabular}
\end{table}

\subsection{Ablation Study}

We perform the ablation experiments focusing on both single-modal (audio) and multi-modal (audio-video) tasks. The evaluation is based on the top-1 accuracy.

\begin{itemize}
    \item \textbf{No Confusion Loss:} The model is trained using only the contrastive loss functions, without confusion loss.
    \item \textbf{No Dynamic Weighting:} The confusion loss is treated with a static weight throughout the training process.
    \item \textbf{No Contrastive Learning Framework:} The model is trained without considering the relationships between audio and video features via contrastive learning.
    \item \textbf{No Refinement of Cluster:} The model does not iteratively optimize the clustering results during the self-supervised learning process.
    \item \textbf{No Cluster-guided:} The model only considers coarse-grained feature alignment of audio and video only in audioset during the self-supervised phase, without taking fine-grained categories into account.
    
\end{itemize}

\begin{table}[htbp]
\centering
\caption{Performance Comparison of Different Models on VGGSound Dataset}
\label{tab:xiaorong}
\begin{tabular}{lcc}
\hline
\multirow{2}{*}{\textbf{Method}} 
  & \multicolumn{2}{c}{\textbf{VGGSound (Acc.\%)}} \\
\cline{2-3}
  & \textbf{A} & \textbf{A-V} \\
\hline
DICCAE                        & \textbf{60.1}                   & \textbf{65.5}                   \\
w/o $\mathcal{L}_{CL}$        & 58.2\scriptsize{$\downarrow 1.9$} & 63.1\scriptsize{$\downarrow 2.4$} \\
w/o $\mathcal{M}_{ij}$        & 59.6\scriptsize{$\downarrow 0.5$} & 64.1\scriptsize{$\downarrow 1.4$} \\
w/o $\mathcal{L}_{\mathrm{InfoNCE}}$ & 56.7\scriptsize{$\downarrow 3.4$} & 61.6\scriptsize{$\downarrow 3.9$} \\
w/o Refinement & 58.4\scriptsize{$\downarrow 1.7$} & 63.3\scriptsize{$\downarrow 2.2$} \\
w/o K-means & 57.7\scriptsize{$\downarrow 2.4$} & 62.2\scriptsize{$\downarrow 3.3$} \\
\hline
\end{tabular}
\end{table}

From Table \ref{tab:xiaorong}, it is evident that removing the confusion loss results in a notable performance drop, with a reduction of 1.9\% in audio-only accuracy and 2.4\% in the audio-visual task. This highlights the critical role of capturing and minimizing inter-class confusion during training.

Disabling dynamic weighting leads to a smaller performance degradation, with a 0.5\% reduction in audio accuracy and 1.4\% in the audio-visual task. This suggests that while dynamic weighting is beneficial.

The most significant performance degradation occurs when the contrastive learning framework is removed, with a 3.4\% decrease in audio accuracy and a 4.2\% drop in audio-visual accuracy. This validates the effectiveness of the proposed training framework.

When the No Refinement of Cluster condition is applied, the model shows a 1.7\% drop in audio accuracy and a 2.2\% reduction in audio-visual accuracy. This indicates that the cognitive and learning process of confusing categories requires progressive adjustments.

The No Cluster-guided condition, where only coarse-grained feature alignment is considered on self-supervised phase, results in a 2.4\% decrease in audio accuracy and a 3.3\% drop in the audio-visual task. This demonstrates that fine-grained alignment of audio-video features is essential even at the early stages of cognitive development.

%% file: sec/5_conclusion.tex
\section{Conclusion}

In this paper, we propose the Dynamic Inter-Class Confusion-Aware Encoder (\textit{DICCAE}), a novel approach for improving human activity recognition by addressing the challenge of category confusion in multimodal audio-video tasks. By dynamically adjusting the confusion loss based on inter-class confusion degrees, \textit{DICCAE} effectively enhances the model's ability to distinguish between similar activities. Additionally, we introduce a cluster-guided self-supervised pre-training strategy to address the scarcity of labeled audio-video data, enabling fine-grained alignment of audio-video features. The proposed method sets a strong foundation for future research in multimodal activity recognition and can be extended to other complex spatio-temporal tasks.

%% file: main.bbl
\begin{thebibliography}{31}
\providecommand{\natexlab}[1]{#1}
\providecommand{\url}[1]{\texttt{#1}}
\expandafter\ifx\csname urlstyle\endcsname\relax
  \providecommand{\doi}[1]{doi: #1}\else
  \providecommand{\doi}{doi: \begingroup \urlstyle{rm}\Url}\fi

\bibitem[Akbari et~al.(2021)Akbari, Yuan, Qian, Chuang, Chang, Cui, and Gong]{Akbari21}
Hassan Akbari, Linagzhe Yuan, Rui Qian, Wei-Hong Chuang, Shih-Fu Chang, Yin Cui, and Boqing Gong.
\newblock Vatt: Transformers for multimodal self-supervised learning from raw video, audio and text.
\newblock \emph{NeurIPS}, 34:\penalty0 24206--24221, 2021.

\bibitem[Alayrac et~al.(2022)Alayrac, Donahue, Luc, Miech, Barr, Hasson, and Vinyals]{Alayrac22}
Jean-Baptiste Alayrac, Jeff Donahue, Pauline Luc, Antoine Miech, Iain Barr, Yana Hasson, and Oriol Vinyals.
\newblock Flamingo: a visual language model for few-shot learning.
\newblock \emph{NeurIPS}, 2022.

\bibitem[Baltru{\v{s}}aitis et~al.(2019)Baltru{\v{s}}aitis, Ahuja, and Morency]{Baltrusaitis19}
Tadas Baltru{\v{s}}aitis, Chaitanya Ahuja, and Louis-Philippe Morency.
\newblock Multimodal machine learning: A survey and taxonomy.
\newblock \emph{IEEE TPAMI}, 41\penalty0 (2):\penalty0 423--443, 2019.

\bibitem[Chen et~al.(2020{\natexlab{a}})Chen, Xie, Vedaldi, and Zisserman]{chen2020vggsound}
Honglie Chen, Weidi Xie, Andrea Vedaldi, and Andrew Zisserman.
\newblock Vggsound: A large-scale audio-visual dataset.
\newblock In \emph{ICASSP 2020-2020 IEEE International Conference on Acoustics, Speech and Signal Processing (ICASSP)}, pages 721--725. IEEE, 2020{\natexlab{a}}.

\bibitem[Chen et~al.(2020{\natexlab{b}})Chen, Kornblith, Norouzi, and Hinton]{chen2020simple}
Ting Chen, Simon Kornblith, Mohammad Norouzi, and Geoffrey Hinton.
\newblock A simple framework for contrastive learning of visual representations.
\newblock In \emph{International conference on machine learning}, pages 1597--1607. PmLR, 2020{\natexlab{b}}.

\bibitem[Damen et~al.(2018)Damen, Doughty, Farinella, Fidler, Furnari, Kazakos, Moltisanti, Munro, Perrett, Price, et~al.]{damen2018scaling}
Dima Damen, Hazel Doughty, Giovanni~Maria Farinella, Sanja Fidler, Antonino Furnari, Evangelos Kazakos, Davide Moltisanti, Jonathan Munro, Toby Perrett, Will Price, et~al.
\newblock Scaling egocentric vision: The epic-kitchens dataset.
\newblock In \emph{Proceedings of the European conference on computer vision (ECCV)}, pages 720--736, 2018.

\bibitem[Dosovitskiy(2020)]{Dosovitskiy20}
Alexey Dosovitskiy.
\newblock An image is worth 16x16 words: Transformers for image recognition at scale.
\newblock \emph{arXiv preprint arXiv:201011929}, 2020.

\bibitem[Fayek and Kumar(2020)]{fayek2020large}
Haytham~M Fayek and Anurag Kumar.
\newblock Large scale audiovisual learning of sounds with weakly labeled data.
\newblock \emph{arXiv preprint arXiv:2006.01595}, 2020.

\bibitem[Gemmeke et~al.(2017)Gemmeke, Ellis, Freedman, Jansen, Lawrence, Moore, Plakal, and Ritter]{gemmeke2017audio}
Jort~F Gemmeke, Daniel~PW Ellis, Dylan Freedman, Aren Jansen, Wade Lawrence, R~Channing Moore, Manoj Plakal, and Marvin Ritter.
\newblock Audio set: An ontology and human-labeled dataset for audio events.
\newblock In \emph{2017 IEEE international conference on acoustics, speech and signal processing (ICASSP)}, pages 776--780. IEEE, 2017.

\bibitem[Georgescu et~al.(2023)Georgescu, Fonseca, Ionescu, et~al.]{Georgescu23}
Mihai-Iulian Georgescu, Eduardo Fonseca, Radu~Tudor Ionescu, et~al.
\newblock Audiovisual masked autoencoders.
\newblock \emph{ICCV}, 2023.

\bibitem[Gong et~al.(2021)Gong, Chung, and Glass]{gong2021ast}
Yuan Gong, Yu-An Chung, and James Glass.
\newblock Ast: Audio spectrogram transformer.
\newblock \emph{arXiv preprint arXiv:2104.01778}, 2021.

\bibitem[Gong et~al.(2022{\natexlab{a}})Gong, Liu, Rouditchenko, and Glass]{gong2022uavm}
Yuan Gong, Alexander~H Liu, Andrew Rouditchenko, and James Glass.
\newblock Uavm: Towards unifying audio and visual models.
\newblock \emph{IEEE Signal Processing Letters}, 29:\penalty0 2437--2441, 2022{\natexlab{a}}.

\bibitem[Gong et~al.(2022{\natexlab{b}})Gong, Rouditchenko, Liu, Harwath, Karlinsky, Kuehne, and Glass]{gong2022contrastive}
Yuan Gong, Andrew Rouditchenko, Alexander~H Liu, David Harwath, Leonid Karlinsky, Hilde Kuehne, and James Glass.
\newblock Contrastive audio-visual masked autoencoder.
\newblock \emph{arXiv preprint arXiv:2210.07839}, 2022{\natexlab{b}}.

\bibitem[He et~al.(2016)He, Zhang, Ren, and Sun]{he2016deep}
Kaiming He, Xiangyu Zhang, Shaoqing Ren, and Jian Sun.
\newblock Deep residual learning for image recognition.
\newblock In \emph{Proceedings of the IEEE conference on computer vision and pattern recognition}, pages 770--778, 2016.

\bibitem[Hsu et~al.(2021)Hsu, Bolte, Tsai, Lakhotia, Salakhutdinov, and Mohamed]{hsu2021hubert}
Wei-Ning Hsu, Benjamin Bolte, Yao-Hung~Hubert Tsai, Kushal Lakhotia, Ruslan Salakhutdinov, and Abdelrahman Mohamed.
\newblock Hubert: Self-supervised speech representation learning by masked prediction of hidden units.
\newblock \emph{IEEE/ACM transactions on audio, speech, and language processing}, 29:\penalty0 3451--3460, 2021.

\bibitem[Jaegle et~al.(2021)Jaegle, Borgeaud, Alayrac, Doersch, Ionescu, Ding, and Vinyals]{Jaegle21}
Andreas Jaegle, Sebastian Borgeaud, Jean-Baptiste Alayrac, Carl Doersch, Catalin Ionescu, David Ding, and Oriol Vinyals.
\newblock Perceiver io: A general architecture for structured inputs \& outputs.
\newblock pages 4651--4664, 2021.

\bibitem[Khosla et~al.(2020)Khosla, Teterwak, Wang, Sarna, Tian, Isola, Maschinot, Liu, and Krishnan]{khosla2020supervised}
Prannay Khosla, Piotr Teterwak, Chen Wang, Aaron Sarna, Yonglong Tian, Phillip Isola, Aaron Maschinot, Ce Liu, and Dilip Krishnan.
\newblock Supervised contrastive learning.
\newblock \emph{Advances in neural information processing systems}, 33:\penalty0 18661--18673, 2020.

\bibitem[Kipf and Welling(2017)]{Kipf17}
Thomas~N. Kipf and Max Welling.
\newblock Semi-supervised classification with graph convolutional networks.
\newblock \emph{ICLR}, 2017.

\bibitem[Liu et~al.(2021)Liu, Lin, Cao, Hu, Wei, Zhang, Lin, and Guo]{liu2021swin}
Ze Liu, Yutong Lin, Yue Cao, Han Hu, Yixuan Wei, Zheng Zhang, Stephen Lin, and Baining Guo.
\newblock Swin transformer: Hierarchical vision transformer using shifted windows.
\newblock In \emph{Proceedings of the IEEE/CVF international conference on computer vision}, pages 10012--10022, 2021.

\bibitem[Nagrani et~al.(2021)Nagrani, Yang, Arnab, Jansen, Schmid, and Sun]{nagrani2021attention}
Arsha Nagrani, Shan Yang, Anurag Arnab, Aren Jansen, Cordelia Schmid, and Chen Sun.
\newblock Attention bottlenecks for multimodal fusion.
\newblock \emph{Advances in neural information processing systems}, 34:\penalty0 14200--14213, 2021.

\bibitem[Pramanik et~al.(2019)Pramanik, Agrawal, and Hussain]{pramanik2019omninet}
Subhojeet Pramanik, Priyanka Agrawal, and Aman Hussain.
\newblock Omninet: A unified architecture for multi-modal multi-task learning.
\newblock \emph{arXiv preprint arXiv:1907.07804}, 2019.

\bibitem[Radford et~al.(2023)Radford, Kim, Xu, et~al.]{Radford23}
Alec Radford, Jong~Wook Kim, Tao Xu, et~al.
\newblock Robust speech recognition via large-scale weak supervision.
\newblock 2023.

\bibitem[Shaikh et~al.(2024)Shaikh, Chai, Islam, and Akhtar]{shaikh2024multimodal}
Muhammad~Bilal Shaikh, Douglas Chai, Syed Mohammed~Shamsul Islam, and Naveed Akhtar.
\newblock Multimodal fusion for audio-image and video action recognition.
\newblock \emph{Neural Computing and Applications}, 36\penalty0 (10):\penalty0 5499--5513, 2024.

\bibitem[Soomro et~al.(2012)Soomro, Zamir, and Shah]{soomro2012ucf101}
Khurram Soomro, Amir~Roshan Zamir, and Mubarak Shah.
\newblock Ucf101: A dataset of 101 human actions classes from videos in the wild.
\newblock \emph{arXiv preprint arXiv:1212.0402}, 2012.

\bibitem[Srivastava and Sharma(2024)]{srivastava2024omnivec}
Siddharth Srivastava and Gaurav Sharma.
\newblock Omnivec: Learning robust representations with cross modal sharing.
\newblock In \emph{Proceedings of the IEEE/CVF winter conference on applications of computer vision}, pages 1236--1248, 2024.

\bibitem[Wang et~al.(2023)Wang, Wang, Lin, et~al.]{Wang23}
Peng Wang, Shuai Wang, Jing Lin, et~al.
\newblock One-peace: Exploring one general representation model toward unlimited modalities.
\newblock \emph{arXiv preprint arXiv:230511172}, 2023.

\bibitem[Wang et~al.(2020)Wang, Tran, and Feiszli]{wang2020makes}
Weiyao Wang, Du Tran, and Matt Feiszli.
\newblock What makes training multi-modal classification networks hard?
\newblock In \emph{Proceedings of the IEEE/CVF conference on computer vision and pattern recognition}, pages 12695--12705, 2020.

\bibitem[Wu et~al.(2016)Wu, Jiang, Wang, Ye, and Xue]{wu2016multi}
Zuxuan Wu, Yu-Gang Jiang, Xi Wang, Hao Ye, and Xiangyang Xue.
\newblock Multi-stream multi-class fusion of deep networks for video classification.
\newblock In \emph{Proceedings of the 24th ACM international conference on Multimedia}, pages 791--800, 2016.

\bibitem[Xu et~al.(2023)Xu, Zhu, and Clifton]{xu2023multimodal}
Peng Xu, Xiatian Zhu, and David~A Clifton.
\newblock Multimodal learning with transformers: A survey.
\newblock \emph{IEEE Transactions on Pattern Analysis and Machine Intelligence}, 45\penalty0 (10):\penalty0 12113--12132, 2023.

\bibitem[Zellers et~al.(2021)Zellers, Lu, Holtzman, Mishra, Peters, and Choi]{Zellers21}
Rowan Zellers, Ximing Lu, Ari Holtzman, Swaroop Mishra, Matthew~E. Peters, and Yejin Choi.
\newblock Merlot: Multimodal neural script knowledge models.
\newblock \emph{NeurIPS}, 2021.

\bibitem[Zhang et~al.(2019)Zhang, Zhai, Xie, Zhan, and Wang]{zhang2019multimodal}
Su-Fang Zhang, Jun-Hai Zhai, Bo-Jun Xie, Yan Zhan, and Xin Wang.
\newblock Multimodal representation learning: Advances, trends and challenges.
\newblock In \emph{2019 International Conference on Machine Learning and Cybernetics (ICMLC)}, pages 1--6. IEEE, 2019.

\end{thebibliography}
